\documentclass[a4paper,fleqn]{cas-dc}
\usepackage[authoryear]{natbib}
\usepackage{tikz}
\usetikzlibrary{matrix}
\usepackage{graphicx}
\usepackage{CJKutf8}
\usepackage{svg} 
\usepackage[noend]{algpseudocode}
\usepackage{algorithmicx,algorithm}
\usepackage{tabularx}
\usepackage{natbib}
\usepackage{amsmath}
\usepackage{color}

\begin{document}
\let\printorcid\relax 
\let\WriteBookmarks\relax
\def\floatpagepagefraction{1}
\def\textpagefraction{.001}

\shortauthors{Hao Wang et~al.} 
\shorttitle{ }
\title [mode = title]{A Two Dimensional Feature Engineering Method for Relation Extraction} 



\author[1]{\textcolor{black}{Hao Wang}}\fnmark[1]
\ead{gs.hw22@gzu.edu.cn}
\author[1]{\textcolor{black}{Yanping Chen}}\corref{cor1}
\ead{ypench@gmail.com}
\author[1]{\textcolor{black}{Weizhe Yang}}
\ead{wzyang.shin@gmail.com}
\author[1]{\textcolor{black}{Yongbin Qin}}
\ead{qhzheng@mail.xjtu.edu.cn}
\author[1]{\textcolor{black}{Ruizhang Huang}}
\ead{rzhuang@gzu.edu.cn}
\address[author1]{Guizhou University, Guiyang, China}


\fntext[fn1]{Huaxi District, Guiyang City, Guizhou Province, 550025, China.}









\cortext[cor1]{Corresponding author}


\begin{abstract}
Transforming a sentence into a two-dimensional (2D) representation (e.g., the table filling) has the ability to unfold a semantic plane, where an element of the plane is a word-pair representation of a sentence which may denote a possible relation representation composed of two named entities. The 2D representation is effective in resolving overlapped relation instances. However, in related works, the representation is directly transformed from a raw input. It is weak to utilize prior knowledge, which is important to support the relation extraction task. In this paper, we propose a two-dimensional feature engineering method in the 2D sentence representation for relation extraction. Our proposed method is evaluated on three public datasets (ACE05 Chinese, ACE05 English, and SanWen) and achieves the state-of-the-art performance. The results indicate that two-dimensional feature engineering can take advantage of a two-dimensional sentence representation and make full use of prior knowledge in traditional feature engineering. Our code is publicly available at \href{https://github.com/Wang-ck123/A-Two-Dimensional-Feature-Engineering-Method-for-Entity-Relation-Extraction}{https://github.com/Wang-ck123/T4RE}.
\end{abstract}


\maketitle

\section{Introduction}
Relation extraction (RE) is a fundamental task in Information Extraction (IE) which provides support for downstream tasks including text summarization (\cite{takase-kiyono-2021-rethinking}), semantic analysis (\cite{raffel2020exploring}; \cite{scaria2023instructabsa}), knowledge graph construction (\cite{ko2021machine}; \cite{chen2019automatic}), and question answering (\cite{zoph2022st}; \cite{clark2020tydi}). Typically, RE is implemented as a classification task, which assigns a label to indicate the type of relations between each pair of entities in a sentence. For example, consider the sentence “My cat has a problem with his paw” where “paw” and “cat” are named entities. The relation between them is “Component-Whole”. For this specific task, neural network approaches with powerful encoder (CNN, BERT, and Variants of BERT) have demonstrated significant performance (\cite{liu-etal-2022-autoregressive}; \cite{wang-etal-2022-deepstruct}; \cite{huang-etal-2022-unified}; \cite{lyu-chen-2021-relation}; \cite{huguet-cabot-navigli-2021-rebel-relation}). Currently, approaches based on large language models (LLM) have also achieved promising results in RE (\cite{wan-etal-2023-gpt}; \cite{bi2024codekgc}; \cite{guo2023retrieval}; \cite{sainz2023gollie}; \cite{zhang2023llmaaa}).

Neural networks have the advantage of learning abstract features from raw input. It is effective to incorporate a wider range of external features (e.g., dependency trees, part-of-speech (POS) tags, and entity types). However, a sentence usually contains several named entities and the relation between two entities is asymmetric. Because all relation instances that overlap in a sentence should be evaluated, correctly distinguishing them depends on the ability of a network to learn an abstract representation, which encodes contextual features and semantic dependencies relevant to a specific entity pair. Related works in RE have shown that feature engineering is effective in making full use of prior knowledge in shallow architectures (\cite{CHEN2015179}; \cite{xu2021feature}). In deep neural works, feature engineering generates synthesized features. They are also helpful in incorporating external knowledge and capturing semantic structures of a relation instance \cite{chen2021neuralized}.

Recently, a new kind of sentence representation, namely two-dimensional (2D) sentence representation, has emerged in the field of IE, e.g., planarized sentence representation \cite{geng2023planarized} and table filling \cite{YoumiMa2022}. Transforming a sentence into a semantic plane can resolve overlapped relation instances. It also provides a uniform semantic representation for learning contextual features and semantic dependencies of a sentence, e.g., the 2D convolutional operation \cite{ma-etal-2022-joint}. In related works, Elements of the semantic plane refer to abstract representations of possible relation instances. They are generated from a raw sentence by concatenating two word (or entity) representations. Current models directly make a prediction based on each element of the semantic plane. The problem is that it is difficult to make full use of prior knowledge which is valuable to support the RE task.

In this paper, we propose a method for the RE task using feature engineering on the 2D sentence representation. We construct explicit feature injection points in the semantic plane to incorporate combined features obtained through feature engineering based on prior knowledge. This allows neural networks to learn semantic dependencies between adjacent spans on the semantic plane while gaining information enhancement from combined features. In addition, we found that previous studies have ignored the implicit association between combined features and named entities when using them. To alleviate this problem, we design a combined feature-aware attention, which employs an attention mechanism to establish the association between entities and combined features, aiming to achieve a deeper understanding of entities. The contributions of this paper include the following:

\begin{enumerate}[(1)]
\item We propose a 2D feature engineering method for RE, that constructs explicit feature injection points in the 2D sentence representation to accommodate our manually constructed features. It can take advantage of a 2D sentence representation and make full use of prior knowledge in traditional feature engineering.
\item We design a combined feature-aware attention to build the association between entities and combined features to gain a deeper understanding of entities.
\item The experiments on two Chinese benchmark datasets and one English benchmark dataset demonstrate the effectiveness of our method. Our model achieves state-of-the-art performance on these public benchmark datasets.
\end{enumerate}

The remaining portion of this paper is constructed as follows: First, Section \ref{s2} presents related works about RE. Second, Section \ref{s4} provides a detailed discussion of the neural network model we proposed based on the two-dimensional feature engineering method. Third, Section \ref{s5} compares the performance with strong baselines, previous studies, and LLM. Fourth, Section\ref{s7} provides a comprehensive analysis of the model. Finally, Section \ref{s6} provides a conclusion.

\section{Related Work}\label{s2}
Related works can be roughly divided into two categories: sentence-level RE and document-level RE. Sentence-level RE tries to identify relations between entities in the same sentence. On the other hand, document-level RE should find all relations in a document. In recent works, supported by deep neural networks, there is a trend in end-to-end models that are effective in avoiding the cascading failure problem in RE. Therefore, in this section, related works are introduced from three aspects. 

\subsection{Sentence-level RE}

Relation extraction is usually formulated as a classification task, aiming to map entity pairs into a pre-defined label space. In recent years, deep neural networks have been widely used in relation extraction due to their powerful learning capabilities. \cite{cohen2020relation} improved the performance by treating the relation extraction task as a span pair prediction problem. \cite{li2023reviewing} proposed a label graph network with a Top-k prediction set, which has the advantage of dealing with long-tail classes. \cite{li2023sequence} pointed out that the seq2seq architecture can learn semantic information and correlation information from relation names. \cite{zhao2019improving} proposed the concept of an entity graph to represent the correlation between entity pairs, using topological information extracted from the entity pair graph to enrich the model’s classification information.

In addition, researchers often add external knowledge to further improve its performance in capturing semantic information, such as dependency trees, combined features, and n-grams. \cite{chen2021relation} pointed out that previous research using dependency trees mostly focused on the dependency connections between words with limited attention paid to exploiting dependency types. In order to reduce the impact of noise in the dependency tree, they proposed a type-aware map memory structure, which makes reliable judgments on dependency information while using the dependency relationships and types between words. \cite{tian-etal-2021-dependency} proposed an A-GCN model, which uses an attention mechanism to obtain different context words in the dependency tree to distinguish the importance of different word dependencies. \cite{chen2021neuralized} proposed a neuralized feature engineering method that uses combined features to enhance neural networks, improving the performance by encoding combined features as distributed representations. \cite{qin2021relation} used n-grams to construct the input graph for the A-GCN model and improved relation extraction performance by weighting different word pairs from the contexts within and across n-grams in the model.

Currently, LLMs have shown great potential to support a variety of downstream NLP tasks. Researchers are exploring the use of LLM for extracting relations. According to \cite{jimenez-gutierrez-etal-2022-thinking}, LLMs perform poorly on RE tasks due to the low incidence of RE in instruction-tuning datasets. QA4RE\citep{zhang2023aligning} introduces a framework that aligns RE tasks with question-answering (QA) tasks, thus enhancing LLM's performance. GPT-RE\citep{wan-etal-2023-gpt} incorporates task-aware representations and enriching demonstrations with reasoning logic to improve the low relevance between entity and relation and the inability to explain input-label mappings. CodeKGC \citep{bi2024codekgc} improves performance by leveraging code's structural knowledge, using schema ware prompts and rationale-enhanced generation.

\subsection{Document-level RE}

Recently, researchers have begun to focus on document-level relation extraction, where relations can cross multiple sentences in a document. Unlike directly making a prediction based on a sentence marked with two entities, in document-level RE, all entities in a document are usually organized as a graph for prediction. There is a serious data imbalance problem and high computational complexity in classifying all entity pairs in a document, making it more challenging than sentence-level RE.

\begin{figure*}
	\centering
            \includegraphics[width=\textwidth]{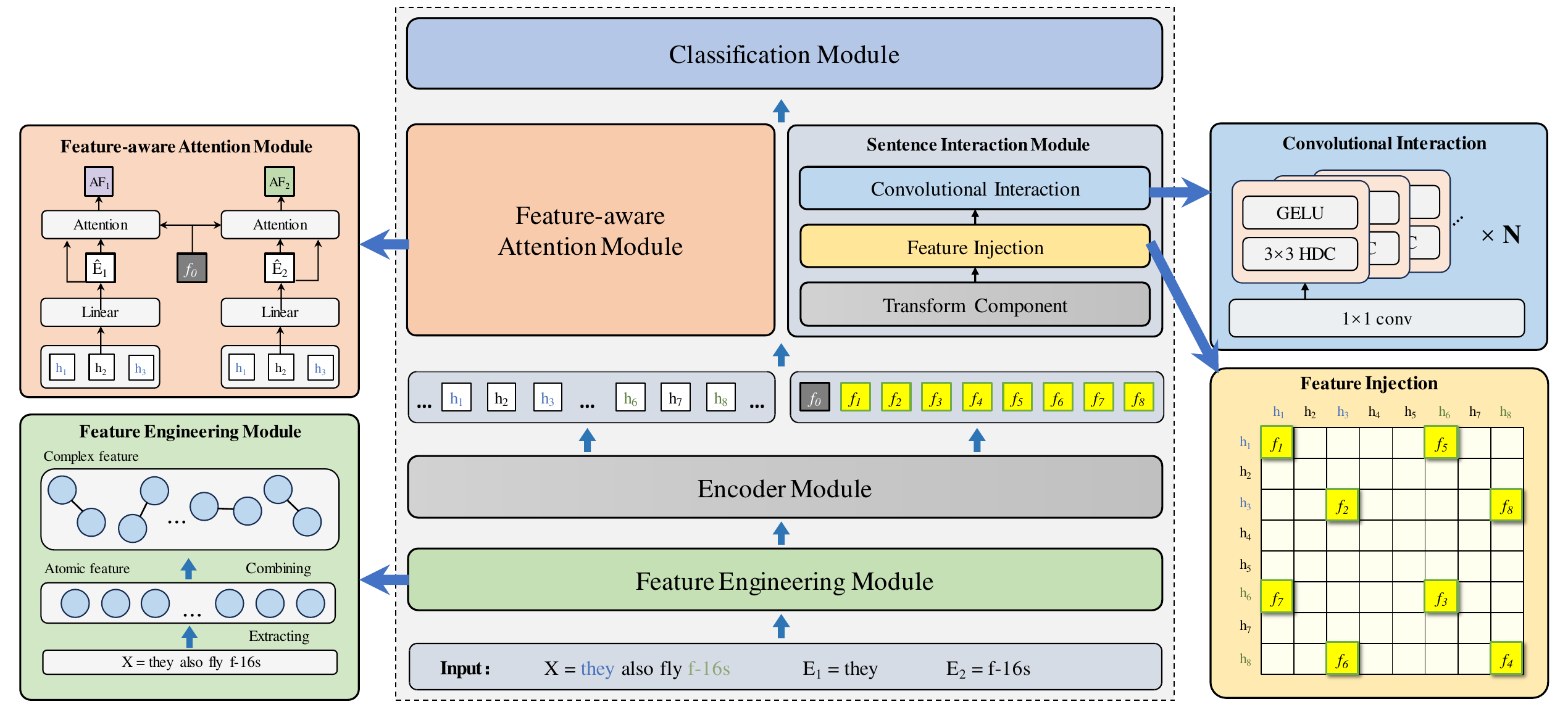}
	\caption{\textrm{{Middle}: The structure of our model illustrated with an example input sentence.}}
	\label{FIG:3}
\end{figure*}

In this aspect, \cite{ma2023dreeam} proposed a memory-efficient method and a self-training strategy to address the high memory consumption and the insufficient data issues in evidence retrieval for document-level relation extraction. \cite{tan2022document} proposed a new semi-supervised framework that improves the class imbalance and the discrepancy between human-annotated data and distant supervision data by using adaptive focal loss and knowledge distillation. \cite{xu2021entity} proposed a method that combines standard self-attention with entity structure, which can use the entity structure information to guide the model for relation extraction. \cite{xie2021eider} proposed an effective evidence enhancement framework.

\subsection{End-to-end RE}

In relation extraction, the target named entities within a sentence are usually pre-annotated. However, in real applications, the task is often implemented on plain texts without annotated named entities. This limitation has sparked interest in end-to-end relation extraction research, which aims to identify named entities from raw texts and extract the relations between them. Currently, end-to-end relation extraction methods can be divided into two types: pipeline methods and joint methods.

Pipeline methods divide end-to-end relation extraction into two independent subtasks: Named Entity Recognition (NER) and RE. For example, \cite{zhong-chen-2021-frustratingly} used cross-sentence information and entity type annotations to enhance the model’s representation learning ability. \cite{ye2021packed} proposed levitated markers, based on entity type markers, which enables the model to obtain the dependency between span pairs through a specific packaging strategy. The main problem with pipeline methods is that it is difficult to learn the semantic interaction between the tasks.

Joint methods aim to utilize the potential connections between the NER, and RE subtasks by sharing parameters, thereby promoting model performance. \cite{miwa2016end} were the first models to use neural networks in end-to-end relation extraction. This method captures word sequences and dependency tree structures by using LSTM. \cite{zheng2017joint} transformed end-to-end relation extraction into a tagging problem. \cite{eberts2019span} proposed a span-based joint method, which shows advantages in identifying overlapping entities. \cite{luan2019general} achieved shared span representation by dynamically constructing a span graph. \cite{li2019entity} treated end-to-end relation extraction as a multi-round question-answering problem. This method can capture hierarchical dependency  well and could easily integrate reinforcement learning.

In addition, LLMs also achieved breakthrough performance in end-to-end RE. Code4UIE\citep{guo2023retrieval} presents a framework for generating codes using Python classes to define schemas and in-context learning to extract structural knowledge from texts. GoLLIE \citep{sainz2023gollie} fine-tuning LLMs to align with annotation guidelines, enabling them to adhere to guidelines out of the box to improve zero-shot performance. 

\section{The Approach}\label{s4}

Before presenting the two dimensional feature engineering method, we first illustrate the architecture of our model. It is shown in Figure \ref{FIG:3}.

This model is divided into five modules: feature engineering module, encoding module, combined feature-aware module, 2D sentence interaction module, and classification. First, feature engineering is applied to generate combined features encoded with prior knowledge. Second, an encoder, composed of a BERT layer and a Bi-LSTM layer, is adopted to capture contextualized word representations from input sentences. Third, a 2D sentence interaction module transforms sequential word representation into a semantic plane. Fourth, a combined feature-aware attention module is built to capture implicit associations between combined features and entities. Finally, the relation type is predicted by a classification module.

\subsection{Feature Engineering Module}

Let $\mathbf{X}=\{x_1,x_2,\ldots, x_N\}$ be a sentence, where $x_i$ and $N$ denote to a word and the length of the sentence. An named entity $\mathrm{e}_i$ ($i{\in}\{1,2\}$) is a substring of $\mathbf{X}$ ($\mathrm{e}_i \in \mathbf{X}$).

In this paper, the smallest granular features of a relation instance are named ``atomic features''. Six types of atomic features are used in our works, including: (1) Entity type, (2) Entity subtype, (3) POS tags of the words on the left of the entity, (4) POS tags of the words on the right of the entity, (5) Relative order of entities\footnote{There are four kinds of entities relative order: [one] means $\mathbf{e}_1$ is in front of $\mathbf{e}_2$; [two] means $\mathbf{e}_2$ is in front of $\mathbf{e}_1$; [three] means $\mathbf{e}_1$ contains $\mathbf{e}_2$; [four] means $\mathbf{e}_2$ contains $\mathbf{e}_1$.}, (6) Head noun of the entity, as shown in Table \ref{tbl1}. 

Among atomic features, the POS tags are manually generated by an external toolkit. The remaining atomic features are provided by the dataset corpus. 

Based on atomic features, feature engineering is applied to generate two types of combined features, namely complex features and entity markers. They are introduced as follows.

The generated complex features, denoted as $\mathbf{CF}_i$ ($1\leq i\leq8$), are listed as follows, where the symbol ${\parallel}$ represents the feature combination operation.
\begin{align}
     &\mathbf{CF}_1 = \mathrm{LeftPosOf(e_1) \  {\parallel}\   TypeOf(e_1)}\notag\\
     &\mathbf{CF}_2 = \mathrm{RightPosOf(e_1) \  {\parallel}\  TypeOf(e_1)}\notag\\
     &\mathbf{CF}_3 = \mathrm{LeftPosOf(e_2) \  {\parallel}\   TypeOf(e_2)}\notag\\
     &\mathbf{CF}_4 = \mathrm{RightPosOf(e_2) \  {\parallel}\   TypeOf(e_2)}\notag\\
     &\mathbf{CF}_5 = \mathrm{TypeOf(e_1) \  {\parallel}\   TypeOf(e_2)}\notag\\
     &\mathbf{CF}_6 = \mathrm{SubTypeOf(e_1) \  {\parallel}\   SubTypeOf(e_2)}\notag\\
     &\mathbf{CF}_7 = \mathrm{HeadOf(e_1) \  {\parallel}\   HeadOf(e_2)}\notag\\
     &\mathbf{CF}_8 = \mathit{\operatorname{OrderOf (e_1, e_2)}}
\end{align}
where $\mathbf{CF}_i$ denotes the i-th complex feature. The complex feature $\mathbf{CF}_8$ is directly generated from the atomic feature (5) in Table \ref{tbl1}.

Entity markers are generated and implanted into the input sentence to indicate entity types and positions. These markers are referred to as $\mathbf{M}^\rho_i$ where $i\in$\ \{1,2\} and $\rho\in \{r,l\}$ represent the ID of entity and positions of markers. For example, $\mathbf{M}^r_2$ means that this entity marker is implanted in the right side of the first entity $e_2$.

These entity markers are defined as structuralized strings taking the form of: \texttt{$\langle \mathrm{T}\_i \rangle$} and \texttt{$\langle /\mathrm{T}\_i \rangle$}, where $\mathrm{T}$ is the type of entity. \texttt{$\langle \mathrm{T}\_i \rangle$} and \texttt{$\langle /\mathrm{T}\_i \rangle$} are the left and right entity markers of entity $e_i$, respectively. For example, if the type of entity $e_2$ is ``PER'' (Person), the two strings ``\texttt{$\langle \mathrm{PER}\_2 \rangle$}'' and ``\texttt{$\langle /\mathrm{PER}\_2 \rangle$}'' are implanted into the left side and right side of entity $e_2$ in the input sentence.

Complex features and entity markers are the output of feature engineering. Because feature engineering is manually designed with prior knowledge, the complex features are effective in encoding the structural information of a specific entity pair. On the other hand, the entity markers enable a deep network to perceive entity boundaries and build contextual dependencies. Therefore, they have been widely used to support relation extraction \cite{chen2021neuralized, CHEN2015179}.

Before encoding complex features and entity markers into a 2D semantic representation, we create two modified inputs $\widehat{\mathbf{X}}$ and $\mathbf{\widehat{X}_f}$ by utilizing features generated from feature engineering. They are represented as:
\[
\begin{aligned}
& \widehat{\mathbf{X}}=\{x_1,\dots, \mathbf{M}^l_1,e_1, \mathbf{M}^r_1,\dots,\mathbf{M}^l_2,e_2,\mathbf{M}^r_2,\dots,x_N\}\\
&\mathbf{\widehat{X}_f}=\{\mathbf{CF}_1,\mathbf{CF}_2,\mathbf{CF}_3,\mathbf{CF}_4,\mathbf{CF}_5,\mathbf{CF}_6,\mathbf{CF}_7,\mathbf{CF}_8\}
\end{aligned}
\]
where $\widehat{\mathbf{X}}$ is formed by implementing entity markers on both sides of two entities in $\mathbf{X}$, and $\mathbf{\widehat{X}_f}$ is consists of complex features arranged in order.

$\widehat{\mathbf{X}}$ and $\mathbf{\widehat{X}_f}$ are fed into the encoding module for generating 2D semantic representations.

\begin{table}[h,width=.9\linewidth,cols=3]
\caption{\textrm{Atomic Features.}}\label{tbl1}
\begin{tabular*}{\tblwidth}{@{} LLLL@{} }
\toprule
\textbf{\textrm{Ref.}} & \textbf{\textrm{Representation}} & \textbf{\textrm{Description}} \\
\midrule
\textrm{(1)} & $\mathit{\operatorname{TypeOf (e_i)}}$ & \textrm{Type of entity i} &  \\
\textrm{(2)} & $\mathit{\operatorname{SubTypeOf (e_i)}}$ & \textrm{Subtype of entity i} &  \\
\textrm{(3)} & $\mathit{\operatorname{LeftPosOf (e_i)}}$ & \textrm{POS tag of the left word}&  \\
\textrm{(4)} & $\mathit{\operatorname{RightPosOf (e_i)}}$ & \textrm{POS tag of the right word} &  \\
\textrm{(5)} & $\mathit{\operatorname{OrderOf (e_1, e_2)}}$ & \textrm{Relative order of entities} &  \\
\textrm{(6)} & $\mathit{\operatorname{HeadOf (e_i)}}$ & \textrm{Head noun of entity i} &  \\
\bottomrule
\end{tabular*}
\end{table}

\subsection{Encoding Module}\label{sec3.2}

The BERT is adopted as the pre-trained language model (PLM) in the encoding module. Because words may be decomposed into multiple pieces, following \citet{liu-etal-2019-gcdt} and \citet{li2022unified}, the embedding of each word is generated by max-pooling corresponding vector pieces of each word. Let $[]$ It can be formalized as:  
\begin{equation}
\mathbf{h}_i=\operatorname{MaxPool}\left(\boldsymbol{v}_{\operatorname{start}(i)}^{(l)}\ldots\boldsymbol{v}_{\mathrm{end}(i)}^{(l)}\right)
\end{equation}
where, $[\boldsymbol{v}_{\operatorname{start}(i)}^{(l)}\ldots\boldsymbol{v}_{\mathrm{end}(i)}^{(l)}]$ are vector pieces obtained from the BERT for the i-th word in an input. $l$ is the layer index. $d$ is the dimension of the hidden layer. Then, $\mathbf{h}_i$ is the i-th word representation in the sentence after max-pooling. 

To further enhance contextual dependencies, following previous studies (\citet{li2022unified}), a Bi-LSTM layer is set to refine the word representation. For $\widehat{\mathbf{X}}$ and $\mathbf{\widehat{X}_f}$, separated Bi-LSTM layers are applied to refine the word representations.

Finally, the final sentence representations of $\widehat{\mathbf{X}}$ and $\mathbf{\widehat{X}_f}$ can be represented as: 
\[
\begin{aligned}
& \widehat{\mathbf{H}}=\{\mathbf{h}_1,\mathbf{h}_2,\ldots,\mathbf{h}_M\} \\
&\mathbf{\widehat{H}_f}=\{\mathbf{f}_0,\mathbf{f}_1,\mathbf{f}_2,\ldots,\mathbf{f}_8\}
\end{aligned}
\]
where $\widehat{\mathbf{H}}\in \mathbb{R}^{M \times d}$, $\mathbf{\widehat{H}_f}\in \mathbb{R}^{9 \times d}$. $M\operatorname{=}N\operatorname{+}4$ is the length of $\widehat{\mathbf{X}}$,  $\mathbf{f}_0$ is the hidden vector corresponding to the special token [CLS]. It is a special token generated by the BERT, which can be used to represent the entire sentence. $\mathbf{f}_1 \sim \mathbf{f}_8$ are the abstract representations of complex features.

\subsection{Sentence Interaction Module}\label{sec3.3}

This module consists of three components: a transform component, a feature injection component, and a convolution interaction component. The transform component is based on a conditional layer normalization (CLN), which transforms a sentence into a semantic plane. The feature injection component injects complex features into the semantic plane. Then, the convolution interaction component, composed of a 4 layer hybrid dilated convolution, performs cross-correlation operations on the semantic plane for capturing the semantic dependency information between words at different distances. The three components are introduced as follows.

\textbf{Transform Component}: following previous studies (e.g., \citet{li2022unified}), a CLN layer is adopted to transform $\widehat{\mathbf{H}}$ into a 2D sentence representation $\mathbf{P}$, where $\mathbf{P}\in\mathbb{R}^{M \times M \times d}$, $\mathbf{P}_{ij}$ is a possible relation representation composed from two word representation $\mathbf{h}_i$ and $\mathbf{h}_j$. The process can be normalized as:
\begin{equation}
\mathbf{P}_{ij}=\operatorname{CLN}\left(\mathbf{h}_i \mathbf{h}_j\right)=\gamma_{ij} \odot\operatorname{Norm}\left(\mathbf{h}_j\right)+\lambda_{ij}
\end{equation}
where $\operatorname{Norm}\left(\cdot\right)$ is a normalization function. $\mathbf{h}_i$ is the condition to generate the gain parameter $\gamma_{ij}=\mathbf{W}_\alpha \mathbf{h}_i+\mathbf{b}_\alpha$. $\lambda_{ij}=\mathbf{W}_\beta \mathbf{h}_i+\mathbf{b}_\beta$ can be seen as the bias of the normalization layer.

\begin{figure}
\centering
\includegraphics[width=0.7\linewidth]{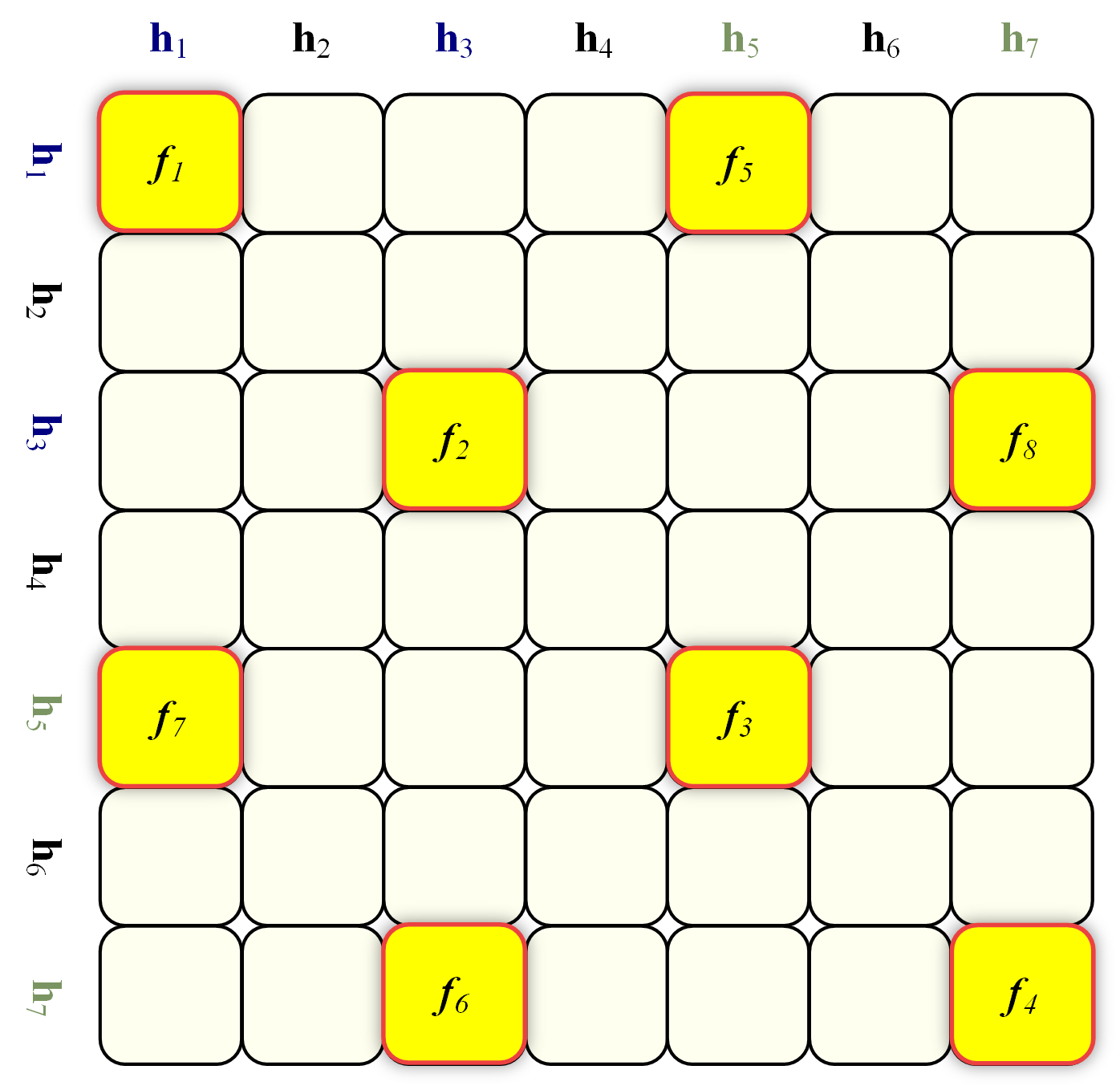}
\caption{\textrm{An example of Feature Injection}}
\label{FIG:2}
\end{figure}

\textbf{Feature Injection (FI)}: the complex feature representations $\mathbf{H_f}$ can be divided into three categories. Features $\mathbf{f}_1$ and $\mathbf{f}_2$ are information relevant to entity $\mathbf{e}_1$. Features $\mathbf{f}_3$ and $\mathbf{f}_4$ are relevant to entity $\mathbf{e}_2$. Features $\mathbf{f}_5 \sim \mathbf{f}_8$ contains information related to both $\mathbf{e}_1$ and $\mathbf{e}_2$. These features are injected into the 2D representation $\mathbf{P}$ according to their characteristics. The strategy to inject features into the 2D representation is demonstrated in Figure \ref{FIG:2}.

In Figure \ref{FIG:2}, $\mathbf{h_1}$, $\mathbf{h_3}$, $\mathbf{h_6}$ and $\mathbf{h_8}$ are representations of Entity markers $\mathbf{M}^l_1$, $\mathbf{M}^r_1$, $\mathbf{M}^l_2$ and $\mathbf{M}^r_2$, respectively. For complex features relevant to a single entity ($\mathbf{f}_1$, $\mathbf{f}_2$, $\mathbf{f}_3$, $\mathbf{f}_4$), they are directly added with the corresponding entity marker representation in the semantic plane. For example, because $\mathbf{f}_1$ denotes to the left POS tag and the entity type of $\mathbf{e}_1$, it is added with the entity marker representation $\mathbf{M}^l_1$ in the diagonal line.

For the complex features relevant to the entity pair ($\mathbf{f}_5$, $\mathbf{f}_6$, $\mathbf{f}_7$, $\mathbf{f}_8$), they are added into the representations composed of two entity boundary representations. For example, $\mathbf{f}_5$ denotes to the entity types of both $\mathbf{e}_1$ and $\mathbf{e}_2$. It is added into the representation composed by $\mathbf{M}^l_1$ and $\mathbf{M}^l_2$ in the upper triangle of the semantic plane. 

After injecting complex features into the semantic plane, a 1$\times$1 2D convolution is adopted to smooth the interaction between features and reduce the dimension of the hidden layer $\mathbf{P}$. The process is expressed by:
\begin{align}
&\mathbf{\ddot{P}}=\operatorname{Conv_{1\times1}} \left(\mathbf{P}\right)
\end{align}
where $\mathbf{\ddot{P}}\in\mathbb{R}^{M \times M \times d^{\prime}}$ is a semantic plane after dimension reduction, $ d^{\prime}$ is the hidden layer dimension of $\mathbf{\ddot{P}}$.

\textbf{Convolutional Interaction}: in this component, a 2D hybrid dilated convolution (HDC) is to capture semantic dependencies between possible relation instances at different distances in $\mathbf{\ddot{P}}$. By using convolutional kernels with different dilation rates in cross-correlation operations, combined features can be leveraged to learn semantic dependencies relevant to specific entity pairs in a sentence.
The HDC component can be formalized as follows:
\begin{align}
&\mathbf{O}_1=\operatorname{Gelu}\!\left(\operatorname{HDC}_{r=1}\left(\mathbf{\ddot{P}}\right)\right)\\
&\mathbf{O}_r=\operatorname{Gelu}\!\left(\operatorname{HDC}_r\left(\mathbf{O}_{r-1}\right)\right)
\end{align}
where $r$ is the dilation rate of the convolution kernel. $\mathbf{O}_r \in \mathbb{R}^{M \times M \times d^{\prime}}$ is the output of the dilated convolution with dilation rate $r$, ``Gelu'' is an activation function. $\operatorname{HDC}_r$ is the convolutional neural network with dilation rate $r$.

In this component, the default value of $r$ is set to 4. The output of each layer is taken out and concatenated in the last dimension to obtain the final output that is $\mathbf{O}=[\mathbf{O}_1;\mathbf{O}_2;\mathbf{O}_3;\mathbf{O}_4]\in \mathbb{R}^{M \times M \times 4d^{\prime}}$.

\subsection{Feature-aware Attention Module}

Because complex features are generated by feature engineering which encodes prior knowledge, they help weight task-relevant features. In this module, based on the multi-head attention mechanism, we design a feature-aware attention (CFA) to capture the interaction between entities and complex features.

Let $[\mathbf{h}_i,\dots,\mathbf{h}_j]$ be the word representations entity $\mathbf{e}_1$ in $\widehat{\mathbf{H}}$, and $[\mathbf{h}_q,\dots,\mathbf{h}_s]$ be the word representations entity $\mathbf{e}_2$ in $\widehat{\mathbf{H}}$. Then, an avg-pooling operation is adopted to obtain vectors $\widehat{\mathbf{e}}_1 \in \mathbb{R}^{d}$ and $\widehat{\mathbf{e}}_2 \in \mathbb{R}^{d}$. They are fed into an activation function and a linear transformation to learn the entity representations, denoted as $\widehat{\mathbf{E}}_1 \in \mathbb{R}^{d}$  and $\widehat{\mathbf{E}}_2 \in \mathbb{R}^{d}$. The process can be formalized as follows:
\begin{align}
&\widehat{\mathbf{E}}_1=\operatorname{Linear}\left(\operatorname{Gelu\left(\frac{1}{j-i+1} \sum_{t=i}^j \mathbf{h}_t\right)}\right)\\
&\widehat{\mathbf{E}}_2=\operatorname{Linear}\left(\operatorname{Gelu\left(\frac{1}{s-q+1} \sum_{t=q}^s \mathbf{h}_t\right)}\right)
\end{align}
where $\widehat{\mathbf{E}}_1 \in \mathbb{R}^{d}$, $\widehat{\mathbf{E}}_2 \in \mathbb{R}^{d}$ and Gelu is an activate function, 

Then, the vector $\mathbf{f}_0$ corresponding to [CLS] from $\mathbf{\widehat{X}_f}$ is used as the query for learning complex feature-aware interactions between entity representations. The multi-head attention mechanism is adopted to capture the association between entities and combined features from a global perspective. The attention mechanism is formalized as follows:
\begin{align}
&\mathbf{AF}_1=\operatorname{MultiHeadAttention}\left(\mathbf{f}_0,\widehat{\mathbf{E}}_1,\widehat{\mathbf{E}}_1\right)\\
&\mathbf{AF}_2=\operatorname{MultiHeadAttention}\left(\mathbf{f}_0,\widehat{\mathbf{E}}_2,\widehat{\mathbf{E}}_2\right)
\end{align}
where, $\mathbf{AF}_1 \in \mathbb{R}^{d}$ and $\mathbf{AF}_2 \in \mathbb{R}^{d}$.

\subsection{Classification Module}

In this module, the classification is based on the output of the 2D interaction module and the feature-aware attention module ($\mathbf{O}$, $\mathbf{AF}_1$ and $\mathbf{AF}_2$). An avg-pooling operation is first implemented on $\mathbf{O}$ to obtain the final sentence representation $\widehat{\mathbf{O}}\in \mathbb{R}^{4d^{\prime}}$. They are concatenated with  $\mathbf{AF}_1$ and $\mathbf{AF}_2$, and fed into an MLP layer and a softmax layer to obtain the final relation distribution $\mathbf{Y}$. The process is represented by:
\begin{align}
\mathbf{Y}=\operatorname{SoftMax}\left(\operatorname{MLP}\left(\mathbf{AF}_1;\mathbf{AF}_2;\widehat{\mathbf{O}}\right)\right)
\end{align}

\begin{table}[h,cols=4]
\caption{\textrm{Statistics of the datasets.}}\label{table2}
\begin{tabular*}{\tblwidth}{@{} lrrr@{} }
\toprule
\textbf{\textrm{Datasets}} & \textbf{\textrm{ACE05\_CN}} & \textbf{\textrm{ACE05\_EN}} & \textbf{\textrm{SanWen}} \\
\midrule
\textrm{Train} &\textrm{85,907}  & \textrm{48,198} & \textrm{13,462} \\
\textrm{Dev} &\textrm{10,738}  & \textrm{11,854} & \textrm{1,347} \\
\textrm{Test} & \textrm{10,739} & \textrm{10,097} &\textrm{1,675}  \\
\textrm{Relation Types} &\textrm{7}  & \textrm{7} & \textrm{9}\\
\bottomrule
\end{tabular*}
\end{table}
\section{Experiment}\label{s5}

Our method is evaluated on three public datasets: 

(1) ACE05 English\footnote{\href{https://catalog.ldc.upenn.edu/LDC2006T06}{https://catalog.ldc.upenn.edu/LDC2006T06}.}(ACE05\_EN), which consists of news, broadcasts, and web blogs. It contains 7 relation types, including no\_relation. We followed the same settings to divide the dataset as in the research of \citet{tian-etal-2021-dependency}, which is 351 documents for training, 80 documents for development, and 80 documents for testing. 

(2) ACE05 Chinese\footnote{\href{https://catalog.ldc.upenn.edu/LDC2006T06}{https://catalog.ldc.upenn.edu/LDC2006T06}.}(ACE05\_CN). The the same as ACE05\_EN, the ACE05\_CN dataset contains the same 7 types of relation labels. We followed the same settings as in the research of \citet{CHEN2015179}, dividing the 107384 instances contained in 633 documents into training, development, and testing at the ratio of 8:1:1. 

(3) SanWen (CLTC), a document-level RE dataset, which consists of 837 Chinese literature articles and includes 9 relation types. We followed the same preprocess as in the research of \citet{wen2018structure} and \citet{xu2017discourse}, which is 695 documents for training, 58 documents for development, and 84 documents for testing. 

Table \ref{table2} shows the statistical information of these three datasets. 

\subsection{Experimental Setup}

In our experiments, we use Pytorch to implement the model and train all the models with AdamW optimizer. We follow previous studies to use NLTK\footnote{\href{https://github.com/nltk/nltk}{https://github.com/nltk/nltk}.} and Jieba\footnote{\href{https://github.com/fxsjy/jieba}{https://github.com/fxsjy/jieba}.} to generate POS tags. To implement word embedding, in the Chinese datasets, we use bert-base-chinese\footnote{\href{https://github.com/huggingface/transformers.}{https://github.com/huggingface/transformers.}} and chinese-roberta-wwm-ext-large as encoders. For the English dataset, we use bert-base-uncased, bert-large-uncased, and roberta-large as encoders. We followed previous research to use the micro-f1 scores for ACE05\_EN and the macro-f1 scores for ACE05\_CN and SanWen. Hyper parameters to implement our model are listed in table \ref{tablehp}.

\begin{table}
\fontfamily{ptm}\selectfont
\caption{\textrm{Hyper parameter settings.}}\label{tablehp}
\begin{tabularx}{\tblwidth}{@{} XX@{} }
\toprule
\textbf{\textrm{Hyper parameter}} & \textbf{\textrm{Value}} \\
\midrule
\textrm{Learning Rate} & [3e-5, 2e-5, 1e-5, 8e-6] \\
\textrm{Warmup Rate} & 0.1 \\
\textrm{Dropout Rate} & [0.5, 0.3] \\
\textrm{HDC nums/Dilation Rate}& [3, 4]\\
\textrm{Batch Size} & [64, 21, 8] \\
\textrm{Trainning epoch} & [20, 12, 10] \\
\bottomrule
\end{tabularx}
\end{table}

\begin{table}
\centering
\caption{\textrm{Comparing with related works}}\label{table31}
\begin{tabularx}{\tblwidth}{@{\extracolsep{\fill}} lll@{}}
\toprule
\textbf{\textrm{Datasets}} &\textbf{\textrm{Models}} & \textbf{\textrm{F1(\%)}}\\
\midrule
\multirow{8}{*}{\textrm{ACE05\_CN}} &\textrm{\cite{che2005automatic}} &\textrm{79.99}\\
&\textrm{\cite{zhang2011developing}} &\textrm{76.13}\\
&\textrm{\cite{8070890}} & \textrm{91.20}\\
&\textrm{\cite{8853305}} &\textrm{90.81}\\
&\textrm{\cite{qin2021entity}(bert-base)} & \textrm{94.94}\\
&\textrm{NFE \citep{chen2021neuralized}(bert-base)} & \textrm{95.87}\\\cline{2-3}
&\textrm{\textbf{Ours}(bert-base)} & \textrm{\textbf{96.75}}\\
&\textrm{\textbf{Ours}(chinese-roberta-wwm-ext-large)} &\textrm{96.42}\\
\midrule
\multirow{8}{*}{\textrm{ACE05\_EN}}&\textrm{DMP\citep{tian-etal-2022-improving} (xlnet-large)}  & \textrm{74.60} \\
&\textrm{NGCN\citep{qin2021relation} (bert-large)} & \textrm{77.72} \\
&\textrm{TaMM\citep{chen2021relation} (bert-large)} & \textrm{78.98} \\
&\textrm{AGCN\citep{tian-etal-2021-dependency} (bert-large)} & \textrm{79.05} \\
&\textrm{CCGN\citep{tian2022combinatory} (bert-large)}& \textrm{79.10}\\
&\textrm{AMT\citep{qin2022enhancing} (bert-large)} & \textrm{79.25} \\\cline{2-3}
&\textrm{\textbf{Ours}(bert-base)}  & \textrm{79.97} \\
&\textrm{\textbf{Ours}(bert-large)} & \textrm{81.42} \\
&\textrm{\textbf{Ours}(roberta-large)} & \textbf{\textrm{82.52}} \\
\midrule
\multirow{7}{*}{\textrm{SanWen}}
&\textrm{ERNIE 2.0\citep{sun2021ernie}}   & \textrm{77.97} \\
&\textrm{DPN\citep{chen2023deep}(bert-base)}    & \textrm{81.25} \\
&\textrm{NFE\citep{chen2021neuralized}(bert-base)}   & \textrm{82.60} \\
&\textrm{ERNIE 3.0\citep{sun2021ernie}}   & \textrm{82.59} \\
&\textrm{ERNIE 3.0 Titan\citep{wang2021ernie}}   & \textrm{82.70} \\\cline{2-3}
&\textrm{\textbf{Ours}(bert-base)}   & \textrm{85.32} \\
&\textrm{\textbf{Ours}(chinese-roberta-wwm-ext-large)}   & \textbf{\textrm{85.53}} \\
\bottomrule
\end{tabularx}
\end{table}

\subsection{Comparing with Related Works}
In this section, our models are compared with previous studies on the three datasets. Our model is implemented with different PLMs. The results are shown in Table \ref{table31}.

In the ACE05\_CN dataset, NFE \citep{chen2021neuralized} was the previous SOTA model. It is a one-dimensional feature engineering method based on a sequential sentence representation, in which combined features are encoded and fed into a deep network for the RE task. Compared with the NFE model, our model adopts a 2D dimensional sentence representation. It has the advantage of representing semantic structures of a sentence, where all related instances are overlapped and share the same contextual features of a sentence.

In the ACE05\_EN dataset, AMT \citep{qin2022enhancing} achieved the best performance. It is a two-training stage multitasking adversarial model that boosts the main relation extractor by explicitly recovering the given named entities. The main problem is that there is a gap between the goals of the two training stages. Compared with the AMT model, the 2D feature engineering can make full use of prior knowledge to support the RE task.

In the SanWen dataset, \textrm{ERNIE 3.0 Titan \citep{wang2021ernie}} was the previous SOTA model. It is a large language model containing 260 billion parameters. This model is self-supervised on a mass of plain texts. It is effective to automatically learn potential knowledge from the raw input. Compared with the ERNIE 3.0 Titan model, our method also achieves better performance. The result indicates that a deep neural network with a vanilla language model tuned with task-relevant prior knowledge and learning objectives is also competitive for a specific downstream NLP task.

 Compared with all related works, our models outperform all previous studies and achieve state-of-the-art performance on the three public benchmark datasets.

\subsection{Compare with Large Language Model}

Currently, large language models (LLMs) have received great attention in the NLP field. LLMs contain a huge number of parameters pretrained in a mass of datasets. LLMs achieve
remarkable performance in various downstream tasks. However, LLMs usually suffer from lower performance. The reason may be that relation directions between entities are asymmetrical and a sentence usually contains several relation instances which share the same context. Because LLMs are generative models, it is difficult for LLMs to distinguish the semantic interaction between named entities.

Because LLM-based models usually focus on few-shot or zero-shot learning which are not effective in making full use of the training data. In this experiment, the GPT-RE model proposed by \cite{wan-etal-2023-gpt} is adopted to compare with our model. The GPT-RE contains a fine-tuned RE model (GPT-RE\_FT), which utilizes fine-tuned relation representations generated by the PURE model \citep{zhong-chen-2021-frustratingly}. To ensure a fair comparison, our model is also tuned with the same data used by the PURE model. It is shown in Table \ref{table20}.

In Table \ref{table20}, ``NULL '' indicates the proportion of no\_relation label in the overall dataset. In this setting, there is a significant increase in the number of no\_relation labels, which requires the model to have a high level of discrimination.

In this experiment, we follow the GPT-RE model to use micro-f1 scores. The GPT-RE uses the ``text-davinci-003'' engine (GPT-3). Our model adopts ``roberta-large'' as our PLM. Table \ref{table33} shows the result of the test set. 

\begin{table}
\fontfamily{ptm}\selectfont\
\caption{\textrm{Statistics of the datasets.}}\label{table20}
\begin{tabular*}{\tblwidth}{@{} lrrrrr@{} }
\toprule
\textbf{\textrm{Datasets}} & Rel. & Train & Dev& Test & NULL \\
\midrule
$\textrm{{ACE05\_EN}}^{*}$ &7 &121,368  & 28,728 & 25,514 &95.98\%\\
\bottomrule
\end{tabular*}
\end{table}

\begin{table}
\centering
\caption{\textrm{Compare with GPT-RE on $\textrm{{ACE05\_EN}}^{*}$}}\label{table33}
\begin{tabular*}{\linewidth}{@{\extracolsep{\fill}} lll@{}}
\toprule
\textbf{\textrm{Models}} & \textbf{\textrm{PLM}} & \textbf{\textrm{F1(\%)}}\\
\midrule
\textrm{GPT-Random} &\textrm{text-davinci-003}  &  \textrm{9.04}\\
\textrm{GPT-Sent} &\textrm{text-davinci-003}  &  \textrm{6.31}\\
\textrm{GPT-RE\_SimCSE} &\textrm{text-davinci-003}  &  \textrm{8.67}\\
\textrm{GPT-RE\_FT} &\textrm{text-davinci-003}  &  \textrm{68.73}\\
\textrm{\textbf{Ours}} &\textrm{roberta-large}  &  \textrm{\textbf{75.66}}\\
\bottomrule
\end{tabular*}
\end{table}

In the above table, ``GPT-Random'' is a random few-shot learning model based on GPT. ``GPT-Sent'' attempts various sentence embedding in retrieval. ``GPT-RE\_SimCSE'' is the model proposed by \cite{wan-etal-2023-gpt}, which uses the SimCSE to compute the similarity between sentence representations. ``GPT-RE\_FT'' denotes to the fine-tuned PURE model \citep{zhong-chen-2021-frustratingly}. This result shows that our model achieves the best performance. Because our model makes use of feature engineering to capture contextual features and semantic dependencies between two named entities, it can effectively identify asymmetric relations between entities.

\begin{table*}
\centering
\caption{\textrm{Ablation study on ACE05\_CN, ACE05\_EN and SanWen}}\label{tbl8}
\fontfamily{ptm}\selectfont
\begin{tabular*}{\linewidth}{@{\extracolsep{\fill}} llllllllll@{}}
\toprule
\multicolumn{1}{l}{\multirow{2}{*}{Model}} & \multicolumn{3}{l}{\textbf{ACE05\_CN}}& \multicolumn{3}{l}{\textbf{ACE05\_EN}}& \multicolumn{3}{l}{\textbf{SanWen}}\\ 
\cmidrule{2-4}\cmidrule{5-7}\cmidrule{8-10}
\multicolumn{1}{l}{}&\textbf{P(\%)}&\textbf{R(\%)}&\textbf{F1(\%)}&\textbf{P(\%)}&\textbf{R(\%)}&\textbf{F1(\%)}&\textbf{P(\%)}&\textbf{R(\%)}&\textbf{F1(\%)}\\ 
\midrule
\multicolumn{1}{l}{full model}&\textbf{96.62}&\textbf{96.90}&\textbf{96.75}&78.33&\textbf{81.67}&\textbf{79.97}&\textbf{86.35}&\textbf{84.71}&\textbf{85.39}\\
\multicolumn{1}{l}{w/o. Bi-LSTM}&94.81&94.23&94.51 &76.14&78.19&77.15&84.47&83.13&83.67\\
\multicolumn{1}{l}{w/o. TI}&95.91&95.71&95.79 &78.60&76.89&77.73&85.24&81.93&83.10 \\
\multicolumn{1}{l}{\ \ \ -w/o. HDC}&95.16&95.08&95.11 &\textbf{80.39}&78.02&79.19&84.54&83.91&83.96\\
\multicolumn{1}{l}{w/o. CFA}&95.54&94.63&95.05 &76.63&76.37&76.50&84.68&84.59&84.60 \\
\multicolumn{1}{l}{\ \ \ -w/o. Linear}&95.87&95.37&95.60 &79.89&79.06&79.48&85.95&83.37&84.31\\
\multicolumn{1}{l}{\ \ \ -w/o. Attention}&93.89&92.69&93.26 &77.82&79.58&78.69&85.35&83.87&84.42\\
\multicolumn{1}{l}{\ \ \ -w/o. Complex feature}&93.60&86.23&89.67 &78.89&73.07&75.87&84.54&82.62&83.40\\
\multicolumn{1}{l}{\ \ \ -w/o. Feature Engineering}&90.14&84.47&87.21 &74.50&70.81&72.61&85.08&82.83&83.67\\
\multicolumn{1}{l}{w/o. Complex feature}&92.68&86.68&89.52 &77.81&75.24&76.50 &85.17&83.67&84.33\\
\multicolumn{1}{l}{w/o. Feature Engineering}&90.03&84.37&87.11 &74.43&71.33&72.85&84.10&82.68&83.28\\
\bottomrule
\end{tabular*}
\end{table*}

\begin{table}
\centering
\fontfamily{ptm}\selectfont\
\caption{\textrm{Case study of our model.}}\label{table34}
\begin{tabular*}{\linewidth}{@{\extracolsep{\fill}} l@{}}
\toprule
\textbf{\textrm{Examples}}\\
\midrule
\textbf{Text A:}\ \textrm{\textcolor{red}{I} spent one summer at the university of \textcolor{blue}{nevada waxing d-}}\\
\textrm{\textcolor{blue}{ormer} to floors and moving furniture in buildings.}\\
\textbf{True Label:} \textrm{org-aff} \\
\textbf{NFE:} \textrm{phys} \\
\textbf{Ours:} \textrm{org-aff} \\
\midrule
\textbf{Text B:}\ \textrm{vivendi universal \textcolor{red}{officials} in the \textcolor{blue}{united states} were not i-}\\
\textrm{mmediately available for comment on friday.}\\
\textbf{True Label:} \textrm{phys} \\
\textbf{NFE:} \textrm{org-aff} \\
\textbf{Ours:} \textrm{phys} \\
\bottomrule
\end{tabular*}
\end{table}

\subsection{Ablation Study}\label{sec:ablation}
In this section, we conduct an ablation study to reveal the contribution of different components in the overall model. The BERT-base is adopted as the encoder in the ablation experiments. ``w/o. component'' means that removing the corresponding component from the full model. ``-w/o. component'' indicates only removing a sub-component from its parent directory. The results are listed in Table \ref{tbl8}, where ``TI'' denotes the 2D interaction module. ``CFA'' denotes the feature-aware attention module.

We first show the influence of removing the Bi-LSTM layer (``w/o. Bi-LSTM'') in the encoding module. The Bi-LSTM is adopted to enhance contextual dependencies between word representations. It can be seen that the performance without Bi-LSTM drops 2.24\%-1.72\% on all datasets. The result indicates that Bi-LSTM can effectively assist CLN in generating a high-quality semantic plane.

After removing the 2D sentence interaction module (``w/o. TSI''), the entity-relevant sentence representation $\widehat{\mathbf{H}}$ is directly concatenated with $\mathbf{AF}_1$ and $\mathbf{AF}_2$ for classification. As shown in table Table \ref{tbl8}, compared to the full model, the performance of `w/o. TSI'' degenerates about 2.24\%-0.96\% on all datasets. This result demonstrates that the 2D sentence representation can effectively support sentence-level relation extraction tasks. 

Removing the 2D hybrid dilated convolution (``-w/o. HDC'') also worsens the final performance. The degeneration indicates that the HDC component is helpful in capturing semantic dependencies between possible relation instances at different distances.

The influence of the complex feature aware attention module is shown in ``w/o. CFA''. Compared to the full model, ``w/o. CFA'' drops about 3.47\%-0.72\% on all datasets. This result demonstrates that CFA effectively captures the implicit association between entities with combined features. To further analyze the influence of the CFA component, adhered sub-components of CFA are respectively removed from the CFA component. It was discovered that removing either the Linear or the Attention alone resulted in significant performance degradation in all datasets than removing the entire CFA, which indicates the importance of utilizing both Linear and Attention in the design of the CFA.

In our work, feature engineering is proposed to generate complex features and combined features ?? for advancing the 2D relation representation. To analyze the influence of feature engineering, we set two scenarios to show the impact of combined features and entity markers on the performance. (1) removing them from the full model, (2) removing them from the CFA module. As shown in Table \ref{tbl8}, it is observed that in both scenarios, the absence of complex features and entity markers results in a significant decline in performance. The results indicate the role of feature engineering in our model framework. Complex features are considered refined entity types, and their absence can negatively impact the model’s performance.

\begin{table*}
\centering
\caption{\textrm{Evaluation of Feature Injection. All result are re-implemented with official open-source code or cited from public papers}.}\label{tbl-2}
\fontfamily{ptm}\selectfont
\begin{tabular*}{\linewidth}{@{\extracolsep{\fill}} llllllllll@{}}
\toprule
\multicolumn{1}{l}{\multirow{2}{*}{\textbf{Model}}} & \multicolumn{3}{l}{\textbf{ACE05\_CN}}& \multicolumn{3}{l}{\textbf{SanWen}}& \multicolumn{3}{l}{\textbf{ACE05\_EN}}\\ \cmidrule{2-4}\cmidrule{5-7}\cmidrule{8-10}
\multicolumn{1}{l}{}&\textbf{P(\%)}&\textbf{R(\%)}&\textbf{F1(\%)}&\textbf{P(\%)}&\textbf{R(\%)}&\textbf{F1(\%)}&\textbf{P(\%)}&\textbf{R(\%)}& \textbf{F1(\%)}\\ \midrule
\  R-bert &81.20&72.97&76.76&70.40&68.63&69.32&79.37&74.54&76.88\\
\  NFE &96.66&95.10&95.87&-&-&82.60&80.57&73.83&77.06\\
\  PURE &88.38&81.53&84.63&78.79&77.42&78.02&78.29&74.89&76.55\\
\  PURE(ALB/CRL) &89.52&83.91&86.52&82.41&82.31&82.31&83.15&76.72&79.80\\
\  TYP Marker &93.57&86.62&89.81&86.05&82.79&84.00&79.85&75.35&77.38\\
\  TYP Marker(CRL/RL) &94.51&86.16&90.04&85.42&85.43&85.40&82.51&79.43&80.69\\
\midrule
\  Ours-AFT&96.01&95.67&95.84&85.42&84.48&84.74&76.07&78.71&77.37\\
\  Ours-BEF&96.51&96.25&96.37&84.66&85.74&85.03&76.93&77.06&77.00\\
\  Ours-FR&95.95&94.59&95.25&85.58&83.58&84.36&77.82&80.45&79.11\\
\midrule
\  Ours&96.62&\textbf{96.90}&\textbf{96.75}&\textbf{86.35}&84.71&85.39&78.33&\textbf{81.67}&79.97\\
\  Ours(BL)&-&-&-&-&-&-&82.96&79.93&81.42\\
\  \textbf{Ours(CRL/RL)}&\textbf{96.70}&96.18&96.42&85.82&\textbf{86.01}&\textbf{85.53}&\textbf{83.66}&81.41&\textbf{82.52}\\
\bottomrule
\end{tabular*}
\end{table*}

\section{Analysis}\label{s7}
\subsection{Case Study}
NFE is also a feature engineering based model, which utilizes combined features to capture semantic structures of a relation instance \citep{chen2021neuralized}. The main difference is that NFE is implemented on a sequential sentence representation. In this section, a case study is conducted to compare our model with NFE, as shown in Table \ref{table34}, where named entities $\mathbf{e}_1$ and $\mathbf{e}_2$ in each sentence are highlighted in red color and blue color, respectively.

The results in Table \ref{table34} show that our model can fully consider the contextual information and semantic dependencies in Text A. Even using the same combined features as NFE, our model can accurately determine that ``I'' is employed by ``university of nevada'' rather than indicating a physical location, especially when a preposition ``at'' is occurred in front of the entity ``university of nevada''. In Text B, our model can also correctly infer that the relation between ``officials'' and ``us'' is a physical location relation, not an employment relation. The reason may be that the NFE model easily falls into local semantics around named entities.
\begin{figure}
\includegraphics[width=\linewidth]{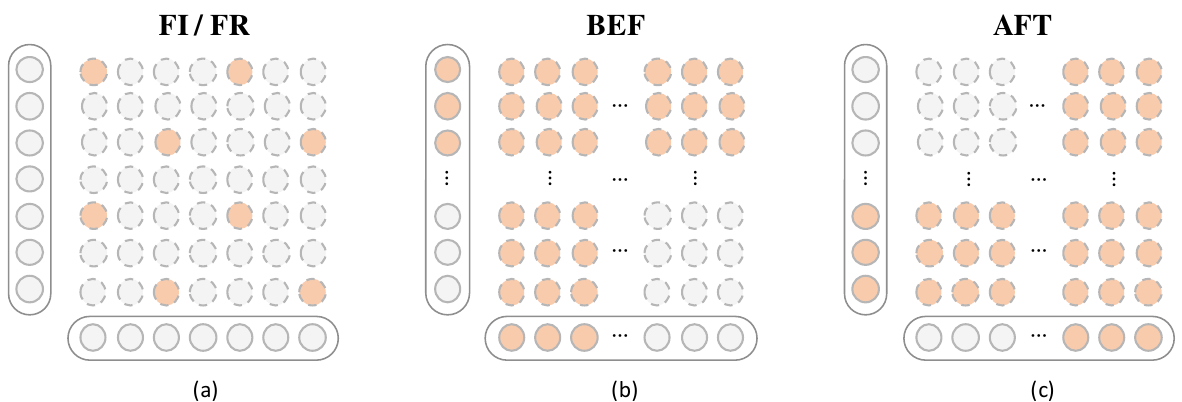}
\caption{\textrm{Different methods of feature injection}}
\label{FIG:2.5}
\end{figure}
On the other hand, our model is more effective in learning long-distance semantic dependencies. 

\subsection{Influence of Feature Injection}
Our strategy to inject features into the 2D representation was presented in Section \ref{sec3.3}, we named this method as \textbf{FI}. In the ablation study (Section \ref{sec:ablation}), feature injection is the most influential component on the final performance (``w/o. complex feature'' and ``w/o. feature engineering'' of Table \ref{tbl8}). In addition to \textbf{FI}, we also designed another three feature injection methods, denoted as ``\textbf{FR}'', ``\textbf{BEF}'' and ``\textbf{AFT}''. (1) \textbf{FI} This is the one we described in Section \ref{sec3.3}. (2) \textbf{FR}  Similar to \textbf{FI} the difference is that \textbf{FR} replaces the values of explicit feature injection points in the semantic plane with complex features. (3) \textbf{BEF} In this method, complex features are concatenated at the beginning of the sentence and input into CLN along with the sentence to be transformed into a semantic plane. (4) \textbf{AFT} In this method, complex features are concatenated at the end of the sentence and input into CLN along with the sentence to be transformed into a semantic plane. As shown in Figure \ref{FIG:2.5}. 

To evaluate the performance of feature injections, they are also compared with four SOTA RE models: R-bert \citep{wu2019enriching}, NFE \citep{chen2021neuralized}, PURE \citep{zhong-chen-2021-frustratingly} and TYP Marker \citep{zhou-chen-2022-improved}. R-bert is a method for RE by enriching the pre-trained BERT model with entity information. NFE generates manually designed combined features to enhance the ability of neural networks to capture structural sentence information in RE. PURE is an effective approach to improve RE performance by introducing entity types. TYP Marker is a powerful RE model that significantly improves the performance of RE through an improved entity representation technique.

For a fair comparison, in this experiment, all results are re-implemented with official open-source code or cited from public papers. Table \ref{tbl-2} shows all results on the same three test datasets, where  BL=Bert-large, ALB=Albert-xxlarge, RL=Roberta-large, CRL=chinese-roberta-wwm-ext-large. The default encoder is Bert-base. Because Bert doesn't have a large version in Chinese, the performance of BL on the ACE05\_CN and SanWen datasets are absent denoted as a ``-'' symbol.

The results show that our model achieves the best performance. It performs on par or even better than TYP Marker on these datasets. The results indicate that feature injection in the semantic plane of prior knowledge is a reliable approach to support the RE task. 

Compared with other models, Ours-BEF and Ours-AFT decrease the performance. Concatenating complex features in the sentence decreases the performance because they are influential when generating the semantic plane. Ours-FR also indicates that ignoring the structure of the semantic plane causes performance degradation. Finally, compared to NFE, which also utilizes combined features, the 2D feature engineering exhibits significantly better performance. Furthermore, the effectiveness of the CFA, which was designed for building associations between entities and combined features, is also demonstrated by the results.

\begin{figure*}
	\centering
        \includegraphics[width=\linewidth]{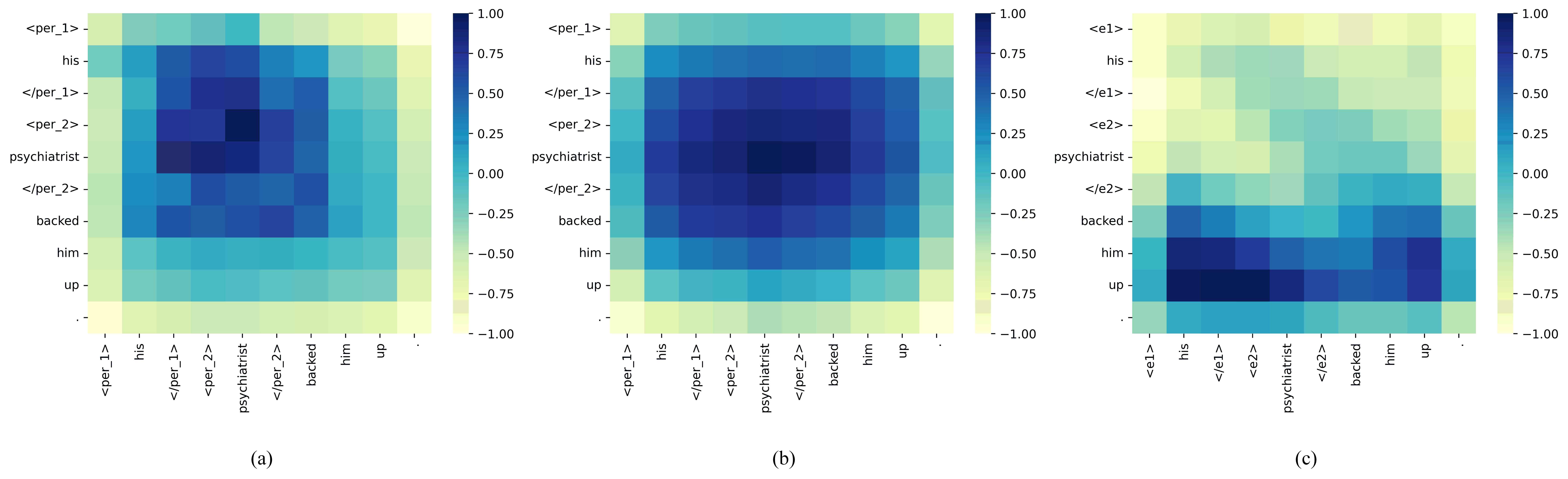}
	\caption{\textrm{Visualization of semantic plane after feature engineering}}
	\label{FIG:4}
\end{figure*}

\subsection{Visualization of the 2D Sentence Representation}

In this section, a visualization is given to show the influence of feature engineering on the 2D sentence representation. The sentence is ``his psychiatrist backed him up'', which contains two named entities (``his'' and ``psychiatrist''). Three scenarios are designed to analyze the effect of feature engineering on the semantic plane. (1) Injecting complex features into the semantic plane. (2) None of the injections in the semantic plane. (3) None entity type in entity markers. We use heat maps to visualize the semantic plane after passing through the 2D sentence interaction module in these three scenarios, as shown in Figure \ref{FIG:4}.

It can be observed that, in Figure \ref{FIG:4} (a), injecting complex features into the semantic plane helps construct the contextual features and semantic dependencies between two named entities. Compared to in Figure \ref{FIG:4} (b), the absence of complex features leads to an even distribution, which is unstable and easily results in a false positive error.

In Figure \ref{FIG:4} (c), when using the $\langle$e\_1$\rangle$ and $\langle$e\_2$\rangle$ tags, which do not carry any type information, the model fails to perceive the entity pair that is difficult to learn task-relevant information from the semantic plane. Compared with Figure \ref{FIG:4} (a), after complex features where injected into the semantic plane, the model is influenced by more granular entity type information in complex features, which helps it capture the dependency information between adjacent spans more accurately. This results in the captured information being more focused on specific words in the entity.

\section{Conclusion}\label{s6}

In this paper, we propose a two-dimensional feature engineering method for entity relation extraction. Our approach guides how to make full use of prior knowledge in the 2D sentence representations. The 2D feature engineering has two advantages to support the RE task. First, a 2D sentence representation is effective in representing complex semantic structures in a sentence, where several relation instances are overlapped and share the same contextual features in a sentence. Second, based on the 2D sentence representation, feature engineering is effective in capturing contextual features and semantic dependencies relevant to a relation instance. Experiments have shown that 2D feature engineering significantly advances the discriminability of a deep neural network and achieves remarkable performance in all evaluated datasets. In further work, the 2D feature engineering method can be extended to support other NLP tasks and give motivation for advancing deep networks by external knowledge.

\section{Acknowledgement}\label{s6.5}
This work is supported by National Natural Science Foundation of China under Grant No. 62066008 and No. 62066007, National Key R\&D Program of China under Grant No. 2023YFC3304500 and the Major Science and Technology Projects of Guizhou Province under Grant [2024]003.

\bibliographystyle{cas-model2-names}

\bibliography{cas-refs}


\end{document}